# Examples of Artificial Perceptions in Optical Character Recognition and Iris Recognition

C.M. Noaica[1], R. Badea[1], I.M. Motoc[1], C.G. Ghica[2], A.C. Rosoiu[3], N. Popescu-Bodorin[4]

*Abstract*— **This paper assumes the hypothesis that *human learning* is *perception based*, and consequently, the learning process and perceptions should not be represented and investigated independently or modeled in different simulation spaces. In order to keep the analogy between the *artificial* and *human learning*, the former is assumed here as being based on the *artificial perception*. Hence, instead of choosing to apply or develop a Computational Theory of (human) Perceptions, we choose to mirror the *human perceptions* in a numeric (computational) space as *artificial perceptions* and to analyze the interdependence between *artificial learning* and *artificial perception* in the same numeric space, using one of the simplest tools of Artificial Intelligence and Soft Computing, namely the perceptrons. As practical applications, we choose to work around two examples: Optical Character Recognition and Iris Recognition. In both cases a simple Turing test shows that *artificial perceptions* of the difference between two characters and between two irides are fuzzy, whereas the corresponding *human perceptions* are, in fact, crisp.**

## I. INTRODUCTION

IN the last decade, Zadeh pointed out the necessity of introducing a Computational Theory of Perceptions (CTP). More precisely, Zadeh outlined in [21] a computational theory of human perceptions, i.e. a formal theory which should enable software agents to compute (and hopefully, to reason) with human perceptions "*described by propositions drawn from a natural language*" [21] as answers to some questions. The process of describing human perceptions in a natural language sets out a correspondence perceptions-propositions and results in what Zadeh called "*perception-based information*" (PBI). Here in this paper PBI is viewed as a special syntactic-semantic representation space, as a subset of legal syntaxes within a given natural language, which are charged with fuzzy meanings. As an example, from a syntactic point of view, there is nothing fuzzy about the string "young person", whereas the fuzzy meaning that humans usually associate to this string makes it f-granular. By paraphrasing Zadeh, this means that the boundaries of perceived "young person" class are unsharp and the values of age ranges are represented in natural language as fuzzy linguistic labels (for example: "young", "middle aged", "old" etc.), as f-granules of (perceived) age, where, as Zadeh said, "*a granule being a clump of values (points, objects) drawn together by indistinguishability, similarity, proximity, and function*" [21].

[1] IEEE Student Members, Artificial Intelligence & Computational Logic Laboratory, Mathematics & Computer Science Dept., Spiru Haret University, Bucharest, Romania; Email: noaica, badea, motoc [at] irisbiometrics.org.
[2] IEEE Student Member, Artificial Intelligence & Computational Logic Lab., Mathematics & Com-puter Science Dept., Spiru Haret University, Bucharest, Romania), Programmer at Clintelica AB, claudiu.ghica [at] clintelica.com.
[3] Game Tester at UbiSoft Romania, alin-cristian.rosoiu [at] ubisoft.com.
[4] IEEE Member, Lecturer, Artificial Intelligence & Computational Logic Laboratory, Mathematics & Computer Science Dept., Spiru Haret University, Bucharest, Romania; Email: bodorin [at] ieee.org

Zadeh illustrated the fact that perceptions are represented in natural language as propositions by giving the following example [21]: "*it is unlikely that there will be a significant increase in the price of oil in the near future*" (which is further denoted here as $\mathcal{A}$) − an assertion that aggregates together fuzzy linguistic labels (words − from the syntactic point of view, perceptions − from the semantic point of view) as aliases of some *f-granules* of *perceived likelihood* ('is unlikely'), *amplitude of variation* ('significant increase') and *time* ('near future').

However, here we assume that instead of representing a human perception, the assertion $\mathcal{A}$ is a piece of knowledge belonging in an economical theory of oil market. How meaningful is $\mathcal{A}$ (the message) depends at least on the writer (the source of the message), reader (the destination of the message) and on the context in which they communicate (the environment). For example, the credibility and the meaning that the reader could assign to the enounce $\mathcal{A}$ vary dramatically when instead of being asserted by a six years old kid it is announced by a spokesman of British Petroleum. Conversely, even if the most qualified economist of the oil market makes the assertion $\mathcal{A}$, there will be dramatic differences between the ways in which different people of different ages and qualifications would assign a meaning to it. Moreover, the assertion $\mathcal{A}$ could be just a part of a communication strategy whose goal is to manipulate the market players.

Since the meanings given by different people to the assertion $\mathcal{A}$ are far enough from being "*drawn together by indistinguishability, similarity, proximity, and function*" [21], the example given by Zadeh actually illustrates that assigning certain meanings to a human perception represented as a proposition in a natural language could result in very volatile / unstable results. On the other hand, such an operation is very similar with trying to decode a message without knowing how exactly the message was encoded in the first place.

This paper assumes the hypothesis that *human learning* is *perception based*, and consequently, the learning process and perceptions should not be represented and investigated independently or modeled in different simulation spaces. On the other hand, in order to keep the analogy between the *artificial* and *human learning*, the former is assumed here as being based on the *artificial perception*. Hence, instead of choosing to apply or develop a Computational Theory of (human) Perceptions [21], we choose to mirror the *human perceptions* in a numeric (computational) space - as *artificial perceptions*, and to analyze the interdependence between *artificial learning* and *artificial perception* in the same numeric space. An alternative that we do not follow here would be to consider learning process as a topic of Artificial Intelligence and perceptions as a topic of a Computational Theory of Perceptions. This paper analyzes the interdependence between learning and perceptions using one of the simplest tools of Artificial Intelligence and Soft Computing, namely the perceptrons [15]. As practical applications, we choose to work around two examples: Optical Character Recognition (OCR) and Iris Recognition (IR).

### A. Outline

The first question that must be answered here is what exactly do we understand here by "*artificial perception*". The second section of this work discusses this topic.

As it can be seen in [21], Zadeh insists on the ideas that (human) "*perceptions, in general, are both fuzzy and granular or, for short, f-granular*" and that "*in much, perhaps most, of human reasoning and concept formation, the granules are fuzzy*" - i.e. human reasoning and the concepts are f-granular.

However, in the two examples that follow to be presented here, the situation is a little bit different: in both cases a simple Turing test [18] shows that *artificial perceptions* of the difference between two different characters and between two different irides are fuzzy, whereas the corresponding *human perceptions* are, in fact, crisp. Despite being contradictory to Zadeh's beliefs expressed above, this situation comes very naturally, because ultimately, a perceptron emulates the human intelligent behavior through an artificial one, which compared to the original is weakened and imprecise enough. The third and the fourth sections from here aim to illustrate this situation in detail using the practical examples of OCR and IR, respectively. Concluding remarks of this study are presented in the fifth section.

## II. THE ARTIFICIAL PERCEPTION

In their seminal work, McCulloch and Pitts [10] formalized the neural networks as a recursively constructed language of *temporal propositional expressions* [10] (build by complexification rules with elementary proposition such as $N_i(t)$ – i.e. the unit i fires at time t). This means that, naturally, a neural network is fully described if we know its structure and if we know why, when and how its neurons fire. Later, Rosenblatt [15], [16] took two important steps further in Artificial Intelligence (AI). Firstly, he defined the Perceptron as an elementary virtual (simulated / artificial) unit able to encode artificial perceptions in a manner similar to that in which is assumed that the human brain supports visual perception [15]. Secondly, he advanced the field of neural networks from a theoretical study to practical implementations on circuits [16].

### A. Artificial Intelligence vs. Human Intelligence

Somehow paradoxically, the two works of McCulloch, Pitts [10] and Rosenblatt [15], [16], taken together, still drawn the limits in which the neural networks have been modeled and used up to this date. Of course, some improvements and some diversification are visible with respect to the structure (recurrent networks, for example), dynamic (self-organizing maps [9], for example), neuron design (fuzzy [19] and neo-fuzzy neurons [20]) and to the area of application [1], [2], [4], [5], [7], [8]. However, all of these newer developments and variations are much too close, and too tributary to the initial design specified by McCulloch, Pitts and Rosenblatt. Maybe this is the reason why, as J. Copeland said in [6], "*five decades since the inception of AI have brought only very slow progress, and early optimism concerning the attainment of human-level intelligence has given way to an appreciation of the profound difficulty of the problem*". Still, it should be very clear for anybody that there is no such logical thing as criticizing AI in itself. All AI tools are our creations and they have only those limitations that we cannot overcome when we design them. Hence, the problem of programming artificial intelligent agents is recursively depending on itself: the AI tools would not have unwanted/unexpected limitations if someone (or something) could be able to design them intelligently. The lower our level of understanding (our own) intelligence, the greater the limitations of the AI tools that we design. The vicious circle does not break here because, in order to find what intelligence is, we should start knowing/acquiring this concept through as many of its hypostases as possible, but on the other hand, the task of recognizing the hypostases of intelligence is not always simple and successful. Moreover, even that we may recognize a hypostasis of intelligence, there is no guarantee that we could understand how it is produced. Hence ultimately, if we accept that AI is "*the science of making computers do things that require intelligence when done by humans*" [6], we should also accept that the limitations of our AI tools originate in our limited knowledge of ourselves and, by consequence, in our limited capacity of designing them intelligently.

Another problematic issue in the present state of AI is a very well established tendency for overvaluation, whose roots grown right from the beginnings. For example, despite the fact that Rosenblatt introduced the perceptrons as elementary units designed to encode artificial perceptions (as their name suggests), when he approached the "*mathematical analysis of learning in the perceptron*" [15], he involuntarily made an association between *learning* and *perceptrons*. Over time, this association somehow has come to be treated, seen and claimed as it would be a strong bound between learning and perceptrons, a bound whose strength increased in time for no plausible logical reasons, only by frequent use, overvaluation and mistake.

### B. Overvaluation

The belief that *perceptrons learn* is widely spread today in AI community and often treated as an objective fact. The truth is that *human learning* is something much more complex than the process of "*learning in the perceptron*" [15]. The latter is nothing more than an expression denoting a very basic piece of learning, namely a process that encodes (memorizes) experiences in a numeric space. This process results in a collection of numbers called trained memory. However, the process of *human learning* results (among many other things) in texts like this one that we write here, i.e. in well-articulated logical discourses about a certain part of reality perceived by us. Our discourses are pieces of complex knowledge expressed with formal correctness and produced by our brain – a complex system of neural networks able to play cooperative games, able to make massively distributed and massively parallel computations, but unfortunately unable to describe itself. This is why reverse engineering the human brain is one of the most relevant tasks for all AI sub-disciplines of our days and is the only task that could make us hope we will ever succeed to endow a machine with *artificial learning* capabilities.

The discrepancy between expectations and achievements in AI is fueled by over-valuation in the first place. Any objective and well-educated mind knows that what a system could achieve depends on how controllable it is and on what states are observable. In other words, from a formal standpoint, what is achievable on a given system is syntactically correct and semantically relevant (all in all, is formally demonstrable) in the formal language and theory that describe the system. Hence, full understanding of a given system means that our expectations and system behavior perfectly match each other.

The overvaluation occurs especially in the cases in which an observable state is treated as being something that it is not and/or as having properties that it does not have. A special paradigm of *learning* treated in AI is that of supervised learning by exposure to examples. The simplest case assumes that a *perceptron learns* to differentiate between two classes of examples (learns a binary classification). We marked the sequence "*perceptron learns*" because, as this paper follows to show, it is more appropriate to say that the (memory of a) *perceptron encodes* the separation between two linearly separable classes of examples. The process of *learning* is far much complicated (as a routine) and far more spectacular as results than the simple mechanical encoding procedure through which a perceptron memorizes the separation between two classes of examples. Human learning, artificial learning, artificial perception and mechanical encoding actually are four different things in AI: *human learning* is a target behavior, *artificial learning* is a computational simulation of human learning, *artificial perception* is an analogy of *human perception*, learning is perception based and *mechanical encoding* is a procedure that may allow artificial perception. Saying that *mechanical encoding* is the same thing with *artificial perception* or with *artificial learning* is nothing else than overvaluation. Talking about learning without making the differences between *human* and *artificial learning* is also an overvaluation of the latter.

### C. What is the Artificial Perception?

Firstly, let us discuss about the *artificial representation* of *human perception* and to illustrate them on the simple case of a binary classification encoded as a trained memory. Let us consider two classes $C^+$ and $C^-$ of linearly separable n-dimensional positive and negative examples, respectively (as those represented in Fig. 1 ), used as training examples for the artificial neuron (perceptron) described in Fig. 2, where:

$$X = [x_1, x_2, \dots, x_n]$$

is the current example applied to the neuron,

$$W = [w_1, w_2, \dots, w_n]$$

is the synaptic memory, $\theta$ is the threshold and the fire function $Y^\pm(X) = f^\pm(X, W, \theta)$ establishes the instant input-output relation of the neuron:

$$f^\pm(X, W, \theta) = \text{sign}(W.X - \theta), \qquad (1)$$

as a function depending on the instant internal activation of the neuron:

$$h(X, W) = W.X, \qquad (2)$$

and on the threshold $\theta$, where the dot operation in formula (2) signifies the scalar product.

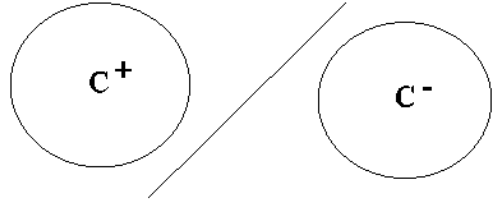

Fig. 1. Graphical representation of two linearly separable classes

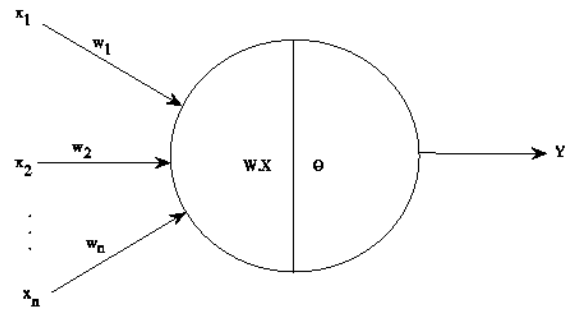

Fig. 2. Graphical representation of a perceptron

With these notations, the (memory of the) neuron is said to be trained if and only if:

$$\forall X \in C^+ \cup C^-, \ Y^\pm(X) = IC^+(X) - IC^-(X), \qquad (3)$$

where $IC^+$ and $IC^-$ are the regular indicator functions (binary membership functions / characteristic functions) of classes $C^+$ and $C^-$, hence:

$$Y^\pm(X) = 1 \text{ iff } X \in C^+ \text{ and } Y^\pm(X) = -1 \text{ iff } X \in C^-. \qquad (4)$$

Hence, if the (memory of the) neuron is trained, then:

$$Y^{\pm}(C^+) = \{+1\} \text{ and } Y^{\pm}(C^-) = \{-1\}. \tag{5}$$

The case from above describes an artificial neuron with bipolar output ($\pm 1$). When the output is binary, the classes may be denoted as $C^1$ and $C^0$, whereas the (memory of the) neuron is said to be trained if and only if:

$$\forall X \in C^0 \cup C^1, Y^{01}(X) = IC^1(X), \tag{6}$$

where $IC^1$ is the indicator function of class $C^1$ and the new instant input-output relation is as follows:

$$f^{01}(X, W, \theta) = logical( f^{\pm}(X, W, \theta) + 1 ) \in \{0,1\}. \tag{7}$$

With these notations, if the (memory of the) neuron is trained, then:

$$Y^{01}(C^1) = \{1\} \text{ and } Y^{01}(C^0) = \{0\}. \tag{8}$$

In the two cases described above, each of the formulae (5) and (8) is an *artificial representation of the human perception* illustrated in Fig. 1, representation written in a first order logico-arithmetic formal language that aggregates constants ($\pm 1$ or 0,1), inputs (X), states / memory instances (W, $\theta$) and outputs (Y) accordingly to the production rules (1)-(8), as appropriate. If we do not wish to know such an artificial representation as a predicate, we still have the possibility to know it as an internal state (as a numeric constant), namely as the trained memory ($W_t$, $\theta_t$). However, interpreting the assertion ($W_t$, $\theta_t$) means to assign certain meanings to (a human perception represented as a) proposition ($W_t$, $\theta_t$) written in a numerical language, a process that could result in very volatile / unstable results if the encoding-decoding rules (1)-(8) are not known, just like in the case of assertion $\mathcal{A}$ discussed in the introduction. On the contrary, knowing the production rules (1)-(8) may allow one to produce a different trained memory $(\hat{W}_t, \hat{\theta}_t)$, which still has the same meaning as ($W_t$, $\theta_t$).

Consequently, the above examples and comments illustrate that human perceptions are encodable (human perceptions can be artificially represented) as trained memory sequences (numeric constants) even by simplified models of the perceptron initially proposed by Rosenblatt [15], [16]. What is new here is the fact that we point out to a second form of artificial representation for the human perceptions, which consists in pieces of formal knowledge (not just in numeric constants), in demonstrable formulae within specific logico-arithmetic formal theories associated to a certain perceptron design.

Regardless their particular type, for the *artificial representations of the human perceptions* to become *artificial perceptions*, it is necessary that the artificial agent who finds and stores them to be self-aware and aware of the meanings that these artificial representations have. Hypothetically, an artificial agent that would actually have *artificial perceptions* should be able to produce propositions like "*I perceive that ...*" and should to be aware of their meanings. Nevertheless, self-awareness and understanding meanings are open problems in AI today. Until the moment when significant progress will have been made on these two directions, the *artificial perception* is just a meaning (and a name) assigned in and by our mind for an *artificial representation of some human perception*. However, we analyze the possibility of developing self-aware software agents based on the basic design of a Cognitive Intelligent Agent given in [14].

## III. Human vs. Artificial Perception of Similarity in OCR

In the particular case analyzed here, the dissimilarity between two hypostases of two different characters is artificially represented as linear separation. For example, when a perceptron instructed to recognize the character 'A' against all the other characters is fully trained, all instances of 'A' from the training set are linearly separated from all instances of all the other characters by a hyperplane whose parameters form the trained memory W. Hence, the difference D between the minimum activation computed for the positive examples and the maximum activation obtained for the negative examples is an artificial perception of the dissimilarity (separation) between the two classes. Let $X_{min}^+$ be the positive example that realizes the minimum activation and let $X_{max}^-$ be the negative example that realizes the maximum activation. With this notations, the number $D/\|W\|$ is the distance between two hyperplanes orthogonal to W, one containing $X_{min}^+$ and the other containing $X_{max}^-$. Hence, the number $d'=D/\|W\|$ is also an artificial perception of the dissimilarity between the two classes (namely: $C^+$ containing 'A' instances, and $C^-$ containing instances of other characters).

On the other hand, the distance between the two sets $C^+$ and $C^-$,

$$d(C^+,C^-)=\min(\{d(X, Y)| \ X \in C^+, \ Y \in C^-\}) \tag{9}$$

is the most objective expression of the dissimilarity between the two classes $C^+$ and $C^-$. The question is how accurate is the artificial perception d' compared to actual distance d.

Let $X_{max}^+$ be the positive example that realizes the maximum activation and let $X_{min}^-$ be the negative example that realizes the minimum activation. Then the number $|W.X_{max}^+ - W.X_{min}^+| \ / \ \|W\|$ is an artificial perception for the diameter of the class $C^+$, whereas the number $|W.X_{max}^- - W.X_{min}^-| \ / \ \|W\|$ is an artificial perception for the diameter of $C^-$.

All in all, the trained memory W sets up an artificial perception (a geometrical view/perspective) that imprecisely encodes the diameter of $C^-$, the distance from $C^-$ to $C^+$ and the diameter of $C^+$ as the numbers $|W.X_{max}^- - W.X_{min}^-| \ / \ \|W\|$, $d(C^+, C^-)$ and $|W.X_{max}^+ - W.X_{min}^+|/\|W\|$. This situation allows us to establish an artificial 3D geometrical conventional representation of the two classes and of the perceived separation between them. The comparison between the ratio d'/d and the real unit tells us when the artificial perception of the separation between the two classes of characters is objective or maximally cointensive with the reality (d'/d=1), fuzzy undervaluated (d'/d<1), or fuzzy overvaluated (d/d'<1). This is why we sustain that even in the classic perceptron the artificial perception / learning is fuzzy (in terms of results). Besides, the update procedure through which the memory of the perceptron changes can be a part of a fuzzy if-then Sugeno rule [17] also. For example, let us consider the following class of if-then linguistic control rules:

IF:
$C^-$ and $C^+$ are two separable classes of examples
And
W is a memory that must be trained
And
N is an *well-chosen* maximal number of epochs
And
R is a *well-chosen* real function having the abscise as asymptote at $+\infty$
And
SP is an *well-chosen* procedure of selecting four examples on which the memory W is trained during an epoch

THEN:
the update rule $(W_{n+1}, t_{n+1}) = (W_n, t_n) + U(SP(n), n)$ converges *rapidly enough* to a *sufficiently well-trained* memory W .

instantiated as follows:

- $C^+$ contains 34 'A' instances, each of them memorized as 8-bit unsigned integer matrices of dimension 16x16, and $C^-$ contains 34 instances for each of the other characters, memorized in the same manner;
- W is a memory randomly initialized;
- N is 1000;
- $R(n) = n*\left(\log_2(n)\right)/2^n$;
- the selection procedure SP return the first two positive examples currently producing the smallest activations $\left(X_{min}^{1+}, X_{min}^{2+}\right)$ and the first two negative examples currently producing the greatest activations $\left(X_{max}^{1-}, X_{max}^{2-}\right)$;
- the update rule assumes that:

$$W_{n+1} = W_n + \left(X_{min}^{1+} + X_{min}^{2+}\right) *R(n), \ t_{n+1} = t_n - \sqrt{\|W_n\|}$$
$$W_{n+1} = W_n - \left(X_{max}^{1-} + X_{max}^{2-}\right) *R(n), \ t_{n+1} = t_n + \sqrt{\|W_n\|}$$

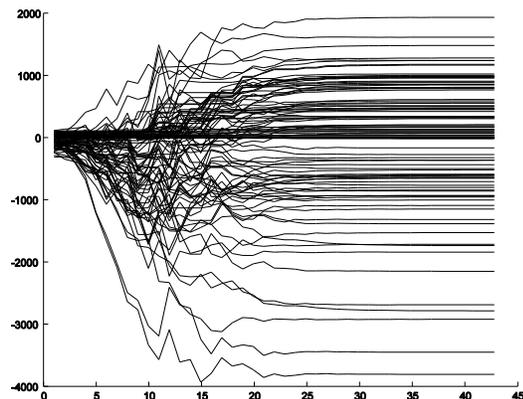

Fig. 3. Convergence of all synaptic memory components along the increasing number of epochs

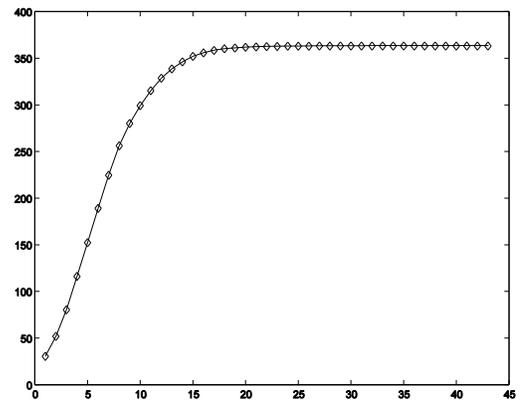

Fig. 4. Convergence of the neuronal threshold along the increasing number of epochs

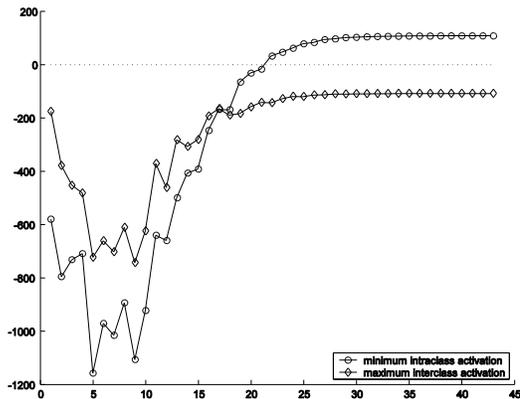

Fig. 5. Convergence of the minimum intraclass activation and maximum interclass activation, along the increasing number of epochs

An implementation of the above fuzzy training rule produced the results presented in figures 3-5 after 43 epochs. The only problem is that the distance between the classes $C^-$ and $C^+$ as it is artificially perceived by the neuron is approximately d'=216, whereas the actual distance between the two classes is approximately d=730, hence the artificial perception of the separation between the two classes of characters is undervaluated (i.e. d'/d<1). The perceived diameters of the classes $C^+$ and $C^-$ are 758 and 380, whereas the actual diameters of the two classes are 2019 and 3060, respectively.

However, accordingly to a very simple Turing test [18], the human decisions on recognizing characters are binary and have nothing to do with the numerical representations mentioned above. This is an example in which the human perception is actually crisp, whereas the artificial perception is actually fuzzy (partial, imprecise).

## IV. HUMAN VS. ARTIFICIAL PERCEPTION OF SIMILARITY IN IRIS RECOGNITION

The most popular way of comparing two binary iris codes is to compute the Hamming distance or the Hamming similarity for the two codes. In this case, the similarity score is a fuzzy value within [0,1]. However, as seen in [13], Hamming distance corresponds to an artificial perception encoded as a synaptic memory referred to as an untrained discriminant direction (see the formulae 1-4 in [13]).

In iris recognition, the training of the discriminant directions aims to diminish the confusion between the fuzzy intervals that underlay the biometric decisions (see Fig.1 from [11], and Fig. 4-5 from [3], for example).

An incipient stage of training the discriminant directions would mean that a narrow safety band (Fig. 2-3 from [13]) separates the two classes of imposter and genuine scores. An advanced stage of training the discriminant directions means that a comfortably wide safety band (Fig. 4-5 from [13]) separates the two classes of genuine and imposter scores. Nevertheless, as seen in Fig. 2-5 from [13], the biometric decisions corresponds to a fuzzy partitioning of the [0,1] interval, regardless the fact that the discriminant directions are trained (sufficiently) or not, i.e. in iris recognition, the artificial perception of the similarity between individuals is fuzzy.

On the contrary, accordingly to a very simple Turing test, the human decisions on recognizing irides are binary (as seen in Fig. 1.a [12]). This is another example in which the human perception is actually crisp, whereas the artificial perception is actually fuzzy (partial, imprecise).

## V. Conclusion

Zadeh insisted on the ideas that (human) "*perceptions, in general, are both fuzzy and granular or, for short, f-granular*" and that "*in much, perhaps most, of human reasoning and concept formation, the granules are fuzzy*". According to this point of view, human reasoning, human concepts and human perceptions are f-granular.

On the contrary, in the two practical examples given here (Optical Character Recognition and Iris Recognition) in order to illustrate the concept of artificial perception, the situation is a little bit different. In both cases a simple Turing test shows that *artificial perceptions* of the dissimilarity between two different characters and between two different irides are fuzzy, whereas the corresponding *human perceptions* are, in fact, crisp. Despite being contradictory to Zadeh's belief expressed above, this fact comes very naturally, because ultimately, a perceptron emulates the human intelligent behavior through an artificial one, which compared to the original is weakened and imprecise enough.

## Acknowledgment

We would like to thank *UbiSoft Romania* and *Spiru Haret University, Romania* for sponsoring this paper and for supporting the research efforts assumed by the team of IEEE Student Members working in the Artificial Intelligence & Computational Logic Laboratory http://fmi.spiruharet.ro/bodorin/aicl/ .